\documentclass{article}
\usepackage{arxiv}
\usepackage[utf8]{inputenc}
\usepackage[T1]{fontenc}
\usepackage{hyperref}
\usepackage{url}
\usepackage{booktabs}
\usepackage{amsmath,amssymb}
\usepackage{nicefrac}
\usepackage{microtype}
\usepackage{graphicx}
\usepackage{caption}
\usepackage{subcaption}
\usepackage[numbers, authoryear]{natbib}
\usepackage{multirow}
\usepackage{array}
\graphicspath{{media/}}
\usepackage{mathtools}
\usepackage{tabularx}
\usepackage{longtable}
\usepackage{makecell}
\usepackage{stix}
\DeclareSymbolFont{extraup}{U}{zavm}{m}{n}
\DeclareMathSymbol{\varheart}{\mathalpha}{extraup}{86}
\DeclareMathSymbol{\vardiamond}{\mathalpha}{extraup}{87}

\usepackage{xcolor}
\usepackage{ifthen}
\newboolean{showcomments}
\setboolean{showcomments}{true} 
\newcommand{\yu}[1]{\ifthenelse{\boolean{showcomments}}
{ \textcolor{orange}{(Yu says:  #1)}}{}}
\newcommand{\tian}[1]{\ifthenelse{\boolean{showcomments}}
{ \textcolor{red}{(Tian says:  #1)}}{}}

\newcolumntype{L}[1]{>{\raggedright\let\newline\\\arraybackslash\hspace{0pt}}m{#1}}
\newcolumntype{C}[1]{>{\centering\let\newline\\\arraybackslash\hspace{0pt}}m{#1}}
\newcolumntype{R}[1]{>{\raggedleft\let\newline\\\arraybackslash\hspace{0pt}}m{#1}}

\title{Confidence in Large Language Model Evaluation: \\ A Bayesian Approach to Limited-Sample Challenges}
\author{
  Xiao Xiao, 
  Yu Su,
  Sijing Zhang,
  Zhang Chen,
  Yadong Chen,
 Tian Liu\footnotemark[1] \\[0.5em]  
  \textit{Tencent Hunyuan} \\[0.5em]  
}

\begin{document}
\maketitle
\begin{abstract}
Large language models (LLMs) exhibit probabilistic output characteristics, yet conventional evaluation frameworks rely on deterministic scalar metrics. This study introduces a Bayesian approach for LLM capability assessment that integrates prior knowledge through probabilistic inference, addressing limitations under limited-sample regimes. By treating model capabilities as latent variables and leveraging a curated query set to induce discriminative responses, we formalize model ranking as a Bayesian hypothesis testing problem over mutually exclusive capability intervals. Experimental evaluations with GPT-series models demonstrate that the proposed method achieves superior discrimination compared to conventional evaluation methods. Results indicate that even with reduced sample sizes, the approach maintains statistical robustness while providing actionable insights, such as probabilistic statements about a model’s likelihood of surpassing specific baselines. This work advances LLM evaluation methodologies by bridging Bayesian inference with practical constraints in real-world deployment scenarios.
\end{abstract}

\footnotetext[1]{Corresponding author email: tianxliu@tencent.com}

\section{Introduction}
Contemporary large language models (LLMs) \citep{brown2020language}\citep{ouyang2022training}\citep{radford2018improving}\citep{radford2019language}\citep{zhao2023survey} exhibit probabilistic output characteristics, yet current evaluation systems often employ deterministic methodologies for effectiveness determination. Such assessment mechanisms generate scalar scores as quantitative measures, theoretically grounded in the assumption of a unidimensional latent variable (analogous to the concept of intelligence quotient in humans) that supposedly underlies model capabilities. This quantitative metric then serves as the basis for cross-model comparisons and performance rankings. 

This methodological approach occasionally presents a paradoxical evaluation challenge. Consider two LLM A and B, administering an exam comprising two questions X and Y. Model A correctly answers only X, whereas model B correctly answers only Y, both achieving 50\% accuracy. An uninformed evaluator might equate their performance, perceiving equivalence in competence. However, in most practical settings, questions are administered to multiple LLMs, yielding empirical data on their discriminative capability. For instance, question X is answered correctly by 90\% of models, while question Y is solved by only 20\%. With this context, the evaluation may shift. This raises critical questions about comparative assessment: How should we quantify the likelihood that one model surpasses the other, given their differential performance on a limited number of evaluation tasks?

There are several potential strategies to address this challenge. First, assigning differential score weights to questions X and Y based on their perceived difficulty level, though such scoring remains inherently subjective and fails to account for the statistical properties of LLMs. Second, Pass@N frameworks \citep{chen2021evaluating}\citep{chowdhery2022palm}\citep{touvron2023llama} assess LLMs' upper-bound performance but face implementation challenges due to unspecified decoding strategies (e.g., temperature, top-k, top-p sampling), often requiring tradeoffs between optimizing rare high-quality outputs and ensuring consistent performance. Third, repeated testing with mean and standard error reporting \citep{blackwell2024towards}\citep{miller2024adding} offers a statistical approach but neglects prior knowledge integration.

Educational research employs Rasch models \citep{rasch1993probabilistic}—a specialized item response theory (IRT) tool—to estimate latent student ability through probabilistic measurement. However, determining question difficulty level in IRT often needs distribution assumptions about the test takers capability that is unknown a$\ priori$ in LLMs \citep{polo2024tinybenchmarks}\citep{truong2025reliable}. Elo rating systems—a method originally designed to rank players in competitive games by modeling pairwise outcome probabilities—represent another sophisticated probabilistic approach. By iteratively adjusting scores based on head-to-head comparisons, Elo could theoretically capture nuanced performance trajectories. However, its practical deployment for LLM evaluation is hindered by prohibitive costs, requiring extensive human-labeled comparisons on thousands of queries \citep{chiang2024chatbot}. In practice, constrained sample sizes often limit the ability of typical practitioners or end-users to reliably assess model capabilities through empirical evaluation.

In this study, we present a Bayesian approach that incorporates prior knowledge when assessing a model's latent capabilities. The referenced prior knowledge specifically refers to the response distribution patterns exhibited by diverse LLMs across a curated question dataset. In scenarios where prior information is unavailable, our methodology rigorously adheres to the principle of maximum entropy, a foundational concept in statistical inference that ensures the derivation of statistically optimal and objectively calibrated estimates under conditions of uncertainty.

\section{Bayesian Formulation}
\subsection{Problem Statement}



We begin by establishing the mathematical foundation for model ranking. Let a set of $N$ anchor models be defined as $\mathcal{L} = \{L_1, L_2, \ldots, L_N\}$. For each $L_i \in \mathcal{L}$, its performance on real world tasks is determined by an underlying capability parameter $\theta_i \in \mathbb{R}^+$, which is analogous to human IQ. Without loss of generality, we assume these capabilities follow a monotonically increasing order such that $\theta_1 < \theta_2 < \cdots < \theta_N$. For an out-of-set test model $L_x \notin \mathcal{L}$, our objective is to estimate its probabilistic membership across $N+1$ mutually exclusive ranking intervals: $\theta_x \leq \theta_1$, $\theta_i < \theta_x \leq \theta_{i+1}$ for $i \in \{1,2,\ldots,N-1\}$ and $\theta_N < \theta_x$ through systematic experimentation.

To enable capability discrimination, consider a curated query set $\mathcal{Q}$ comprising $M$ discriminative queries representing real-world operational scenarios. When applied to a model, each query $Q_j$ constitutes a binary random variable ($Q_j \in \{0,1\}$) corresponding to incorrect/correct model responses, and the entire query set constitutes a readout of binary sequence $\{Q_1, Q_2, \ldots, Q_M\}$. The design of the curated query set aims to induce measurable differentiation between the anchor models with different latent capabilities.

\subsection{Model ranking via Bayesian inference}

Given an anchor model set $\mathcal{L}$ and query set $\mathcal{Q}$, the conditional probability of correct response for each anchor model $L_i$ ($i \in \{1,2,\ldots,N\}$) on query $Q_j$ ($j \in \{1,2,\ldots,M\}$) is expressed as $\Pr(Q_j = 1 | L_i)$, which is empirically measurable through controlled experiments. To facilitate subsequent derivations, we introduce two boundary anchor models: $L_0$ with $\theta_0$ and $L_{N+1}$ with $\theta_{N+1}$, where $\forall j \in \{1,2,\ldots,M\}$, $\Pr(Q_j = 1 | L_0) = 0$ and $\Pr(Q_j = 1 | L_{N+1}) = 1$.

Following evaluation of test model $L_x$ on query set $\mathcal{Q}$, we observe binary outcomes $q = \{q_1, q_2, \ldots, q_M\}$. The ranking problem constitutes Bayesian inference over mutually exclusive hypothesis spaces:

\begin{equation}
\Pr(\theta_i < \theta_x \leq \theta_{i+1} | \mathcal{Q} = q) = \frac{\Pr(\mathcal{Q} = q | \theta_i < \theta_x \leq \theta_{i+1}) \Pr(\theta_i < \theta_x \leq \theta_{i+1})}{\Pr(\mathcal{Q}=q)}
\quad \text{for } i \in \{0,1,\ldots,N\}
\label{eq:1}
\end{equation}

\subsection{Parameter Estimation}

Computation of posterior probabilities in Equation ~\ref{eq:1} requires estimation of three components on the right-hand side, denoted as $\Pr_{\text{est}}(\theta_i < \theta_x \leq \theta_{i+1})$, $\Pr_{\text{est}}(\mathcal{Q} = q | \theta_i < \theta_x \leq \theta_{i+1})$, and $\Pr_{\text{est}}(\mathcal{Q})$ for $i \in \{0,1,\ldots,N\}$.

First, following the principle of maximum entropy\citep{jaynes2003probability} without assuming specific $\theta$ distributions:

\begin{equation}
\Pr\nolimits_{\text{est}}(\theta_i < \theta_x \leq \theta_{i+1}) \propto (\theta_{i+1} - \theta_i)
\label{eq:2}
\end{equation}

This ensures uniform prior distribution of $\theta$.

Second, under conditional query independence assumption, the joint likelihood decomposes as:

\begin{equation}
\Pr\nolimits_{\text{est}}(\mathcal{Q} = q | \theta_i < \theta_x \leq \theta_{i+1}) = \prod_{j=1}^M \Pr\nolimits_{\text{est}}(Q_j = q_j | \theta_i < \theta_x \leq \theta_{i+1})
\label{eq:3}
\end{equation}

Conditional probabilities are estimated through:

\begin{equation}
\Pr\nolimits_{\text{est}}(Q_j = q_j | \theta_i < \theta_x \leq \theta_{i+1}) = \frac{\Pr(Q_j = q_j | L_i) + \Pr(Q_j = q_j | L_{i+1})}{2}
\label{eq:4}
\end{equation}

Eq.~\ref{eq:4} again follows the principle of maximum entropy, which postulates that, in the absence of further information about the distribution of $\theta$, the most unbiased assumption is that $\theta$ is uniformly distributed in the interval $(\theta_i,\theta_{i+1}]$. Under this assumption, the linear relationship between $\theta$ and the probability of answer correctness leads to the simplest and most symmetric solution, which is the average of the probabilities at the interval boundaries.

Finally, $\Pr_{\text{est}}(\mathcal{Q}=q)$ serves as a normalization factor ensuring probability conservation:

\begin{equation}
\Pr\nolimits_{\text{est}}(\mathcal{Q}=q) = \sum_{i=0}^N \Pr\nolimits_{\text{est}}(\mathcal{Q} = q | \theta_i < \theta_x \leq \theta_{i+1}) \Pr\nolimits_{\text{est}}(\theta_i < \theta_x \leq \theta_{i+1})
\label{eq:5}
\end{equation}

\section{Experimental Setup}
\subsection{Anchor Model Selection}

In this study, we chose the GPT-series as our anchor models due to their established recognition. Specifically, our anchor models include GPT-3.5 Turbo \citep{openai2023chatgpt}, GPT-4 \citep{openai2023gpt4}, GPT-4o \citep{openai2024gpt4o}, GPT-4.5 \citep{openai2025gpt45}, o1 \citep{openai2024o1}, and o3-mini-high \citep{openai2025o3mini}. The temporal progression of model releases creates an implicit capability hierarchy, with subsequent iterations expecting enhanced performance profiles.

Nevertheless, existing research indicates that chain-of-thought reasoning models typically outperform general-purpose foundation models on benchmark ranking tasks (e.g., Chatbot Arena, SuperCLUE) \citep{chiang2024chatbot}\citep{xu2023superclue}, while distilled variants (i.e., Turbo and mini series) may exhibit performance degradation relative to their full-scale counterparts \citep{gu2023minillm}\citep{wu2024lamini}. To empirically validate the actual capability ordering within this anchor model cohort, we conducted evaluations on our curated query set rather than relying solely on release-order assumptions.

\subsection{Query Set Construction}
To construct the query set, we initially assembled 170 questions by selecting 30 items each from superGPQA \citep{map2025supergpqa}, MMLU-Pro \citep{wang2024mmlupro}, GPQA-Diamond \citep{rein2023gpqa}, MATH \citep{hendrycks2021measuring}, and ZebraLogic \citep{lin2025zebralogic}, along with 10 questions each from KOR-Bench \citep{ma2024korbench} and Procbench \citep{fujisawa2024procbench}. The selection process followed two criteria. First, the initial pool ensured comprehensive task coverage across material science, earth and environmental sciences, engineering and technology, chemistry, medicine, biology, humanities, art, mathematics, physics, and reasoning. Second, based on previous experimental results from model series released in early 2025 (\citep{anthropic2025claude} \citep{google2025gemini} \citep{li2025transmamba} \citep{doubao2025doubao} \citep{qwen2025qwen25}), we implemented discriminative query selection such that no less than one but no more than half of these models could correctly answer each question. 

To emphasize the capability ranking efficacy of our Bayesian evaluation approach even with reduced question sets, we subsequently reduced the query set from 170 to 50 through systematic triage while maintaining source diversity, task diversity, and anchor model performance diversity. This refined set underwent syntactic paraphrasing \citep{dong2023statistical}\citep{yin2024benchmarking} to mitigate recently reported memorization susceptibility in language models \citep{cheng2025survey}\citep{deng2023investigating}\citep{golchin2023data}.

\subsection{Bayesian Ranking}
The evaluation employed five open-source language models---Llama-4-Maverick \citep{meta2025llama4}, DeepSeek-V3-0324 \citep{deepseek2025v3}, DeepSeek-R1 \citep{deepseek2025r1}, QwQ-32B \citep{qwen2025qwq} and Qwen2.5-72B \citep{qwen2025qwen25}---as representative test models. For each test model, evaluations were conducted on the curated query set to derive binary outcome measures. These outcome indicators underwent subsequent mathematical calculations detailed in Section~2 to generate probabilistic ranking scores.

Specifically regarding prior knowledge acquisition, we repeatedly tested each anchor model $L_i$ on each question $Q_j$ for $O=10$ times and recorded the number of successes to calculate the conditional probability $\Pr(Q_j | L_i)$. To ensure numerical stability during calculation, we introduced $\epsilon$-adjustment mechanisms ($\epsilon=0.01$) that apply $\pm 1\%$ boundary modulation when observing extremal probability values ($\Pr(Q_j | L_i) \in \{0,1\}$), thereby preventing division-by-zero errors in subsequent computations. This information was then systematically organized in a capability matrix $\Pi \in \mathbb{R}^{N \times M}$:

$$
\Pi = 
\begin{bmatrix}
P(Q_1 | L_1) & P(Q_2 | L_1) & \cdots & P(Q_M | L_1) \\
\vdots & \vdots & \ddots & \vdots \\
P(Q_1 | L_N) & P(Q_2 | L_N) & \cdots & P(Q_M | L_N)
\end{bmatrix}
$$

The comparative ranking among anchor models was ultimately determined through cumulative success rates aggregated across the entire question set, with $\theta_i$ representing the cumulative success rate for $L_i$. Optimal anchor model-query set pairing produces uniform interval sizes ($\theta_{i+1} - \theta_i$) that ensure consistent discriminatory power across the entire capability spectrum.

Furthermore, we conducted additional experiments by reducing the query set size to 30, 20, 10, and 5 to systematically evaluate the robustness of the proposed methods.

\subsection{Methods comparison}
We compare the Bayesian method with three other commonly employed evaluation methods, namely: simple accuracy reporting, Pass@N reporting, and $\mathrm{mean}\pm\mathrm{std}$ reporting. 

While accuracy reporting requires a single trial of evaluating a test model on the query set, the Pass@N selection method and $\mathrm{mean}\pm\mathrm{std}$ reporting protocol require multiple trials. Therefore, we conducted $O=10$ independent trials on each evaluated model $L_x$ against the query set $\mathcal{Q}$.

In the proposed Bayesian evaluation approach, we integrate multi-trial observations through stochastic modeling of the query $\vec{Q}_j = \{Q_j^{(1)}, Q_j^{(2)}, \ldots, Q_j^{(O)}\}$ as a collection of independent Bernoulli trials, where $O$ represents the total number of trials conducted for query $Q_j$. For the evaluation protocol, the corresponding measurement is defined as $\vec{q}_j = \{q_j^{(1)}, q_j^{(2)}, \ldots, q_j^{(O)}\}$, containing $K$ correct outcomes. This formulation generalizes Eq.~\ref{eq:4} to accommodate multiple trials:

\begin{equation}
\Pr\nolimits_{\text{est}}(\vec{Q}_j = \vec{q}_j | \theta_i < \theta_x \leq \theta_{i+1}) = 
\frac{\Pr\nolimits_{\text{est}}(\vec{Q}_j = \vec{q}_j | L_i) + \Pr\nolimits_{\text{est}}(\vec{Q}_j = \vec{q}_j | L_{i+1})}{2},
\label{eq:6}
\end{equation}

where the likelihood function for each anchor model $L_i$ under the multi-trial paradigm is estimated as:

\begin{equation}
\Pr\nolimits_{\text{est}}(\vec{Q}_j = \vec{q}_j | L_i) =
\binom{O}{K}\Pr(Q_j = 1 | L_i)^K \Pr(Q_j = 0 | L_i)^{(O-K)}
\label{eq:7}
\end{equation}

This generalized formulation explicitly incorporates the multi-trial nature of the evaluation process. Notably, when $O=1$, Equation~\ref{eq:7} analytically reduces to the single-trial probability expression, thereby establishing mathematical consistency between single-trial and multi-trial evaluation regimes.

\section{Results}

\begin{figure}[htbp]
\centering
\includegraphics[width=1\textwidth]{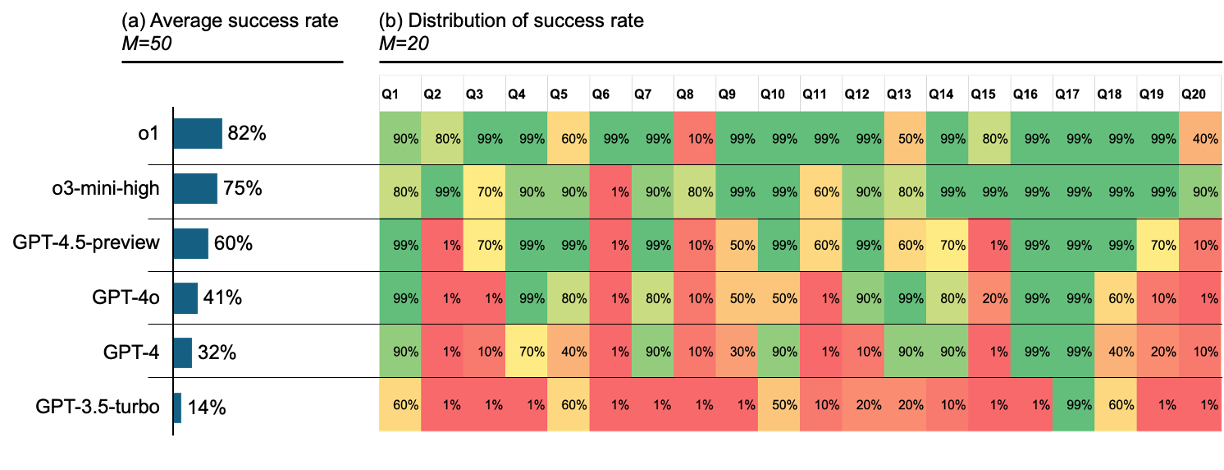}
\caption[Anchor Model Performance]{\textbf{Anchor Model Performance}\\
\vspace{4pt} 
\begin{minipage}{\textwidth}
(a) Success rates of the six anchor models (measured over $O=10$ trials per question) for $M=50$ evaluation questions. \\
(b) The success rate of the first 20 questions are shown here, where the complete success rate distribution is provided in Appendix. Extreme probability values $\{0\%,100\%\}$ were modulated to $\{1\%, 99\%\}$ to ensure numerical stability during subsequent computations.
\end{minipage}
}
\label{fig1}
\end{figure}

The capability matrix was systematically visualized in Figure~\ref{fig1}, where anchor model performance was shown as their respective probability of correctly answering the questions. Anchor models demonstrated quasi-uniform spacing in their cumulative success rate (14\% to 82\%), revealing effective capability differentiation within the query set (complete query set provided in ~\ref{Complete Query Set}). 

Notably, model performance exhibited temporal progression aligned with release timelines (GPT-3.5 turbo: 14\% vs.\ o1: 82\%). However, reasoning-enhanced systems (o1/o3-mini-high) outperformed non-reasoning models in general, and o3-mini-high---a purported distilled variant of o3---showed degraded performance despite its later release relative to o1.

\begin{figure}[p] 
\centering
\captionsetup[subfigure]{labelformat=empty} 
\vspace*{-0.5cm} 
\begin{subfigure}{\textwidth}
  \centering
  \includegraphics[width=0.85\linewidth,height=0.18\textheight,keepaspectratio]{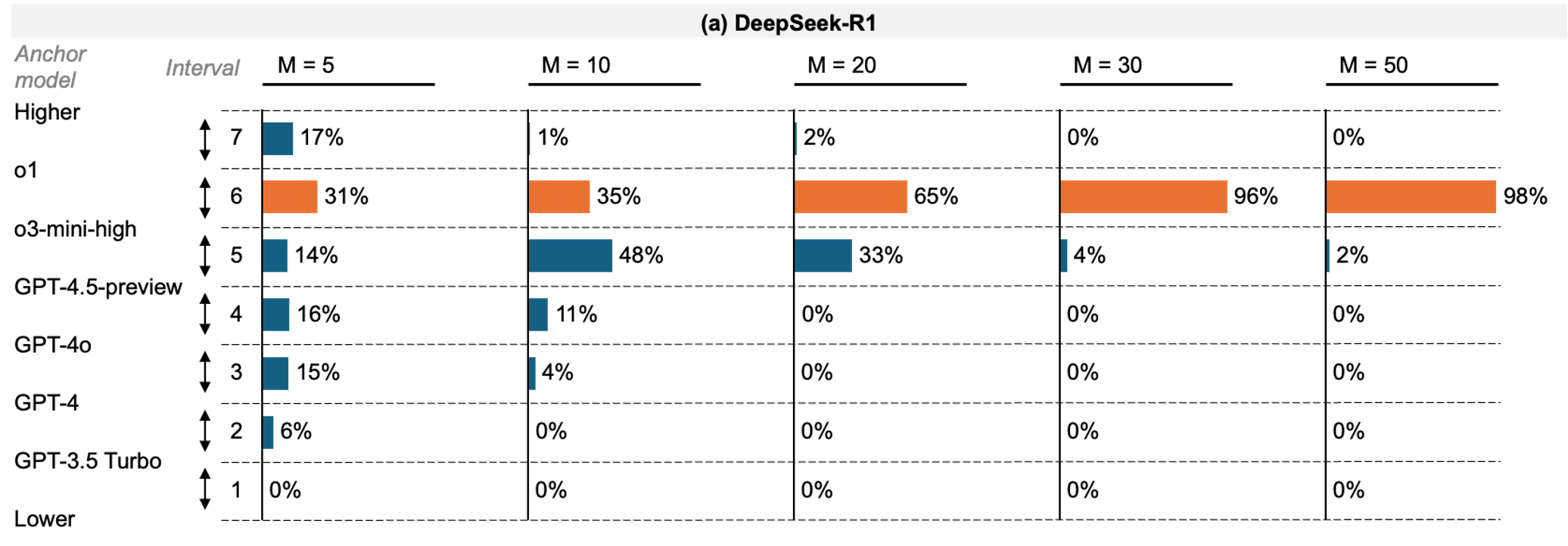}
  \vspace{2pt}
\end{subfigure}

\vspace{0.1cm} 
\begin{subfigure}{\textwidth}
  \centering
  \includegraphics[width=0.85\linewidth,height=0.18\textheight,keepaspectratio]{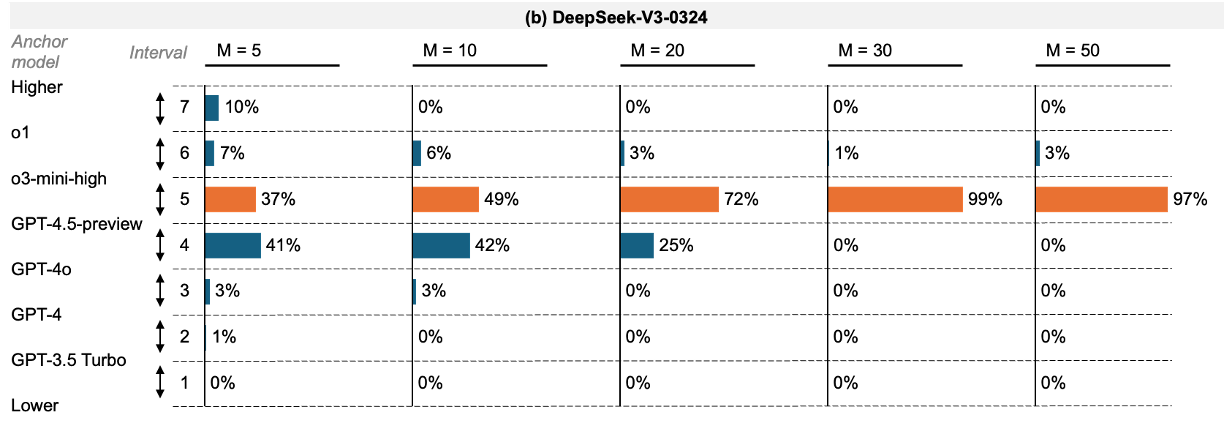}
  \vspace{2pt}
\end{subfigure}

\vspace{0.1cm}
\begin{subfigure}{\textwidth}
  \centering
  \includegraphics[width=0.85\linewidth,height=0.18\textheight,keepaspectratio]{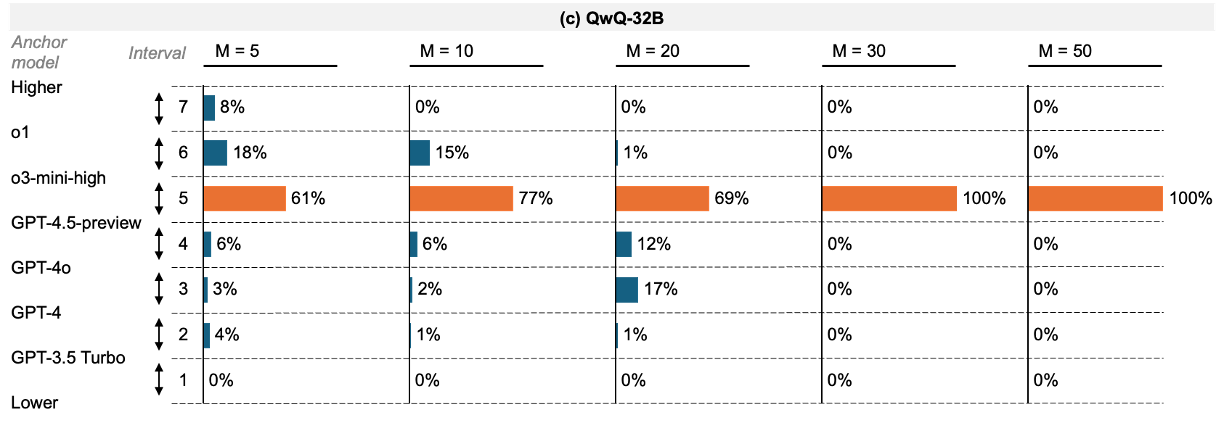}
  \vspace{2pt}
\end{subfigure}

\vspace{0.1cm}
\begin{subfigure}{\textwidth}
  \centering
  \includegraphics[width=0.85\linewidth,height=0.18\textheight,keepaspectratio]{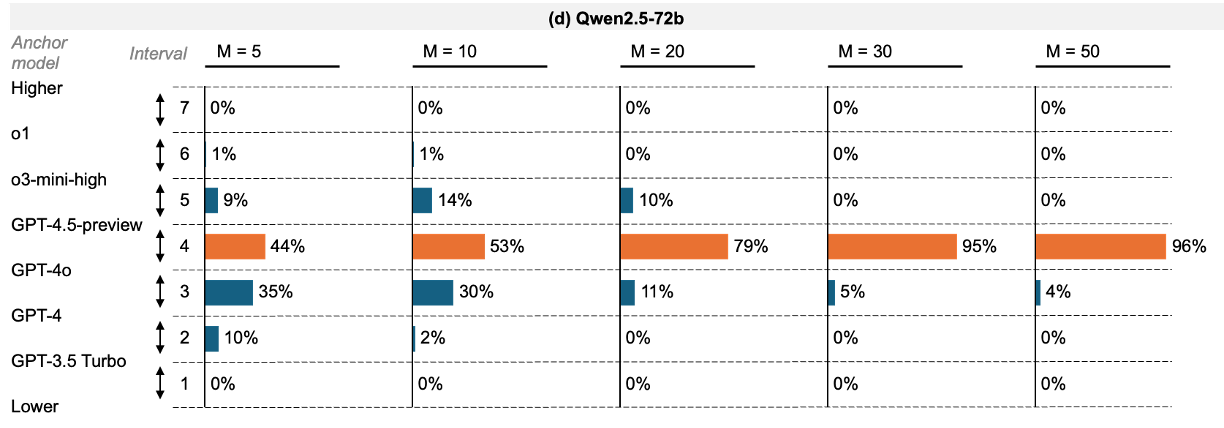}
  \vspace{2pt}
\end{subfigure}

\vspace{0.1cm}
\begin{subfigure}{\textwidth}
  \centering
  \includegraphics[width=0.85\linewidth,height=0.18\textheight,keepaspectratio]{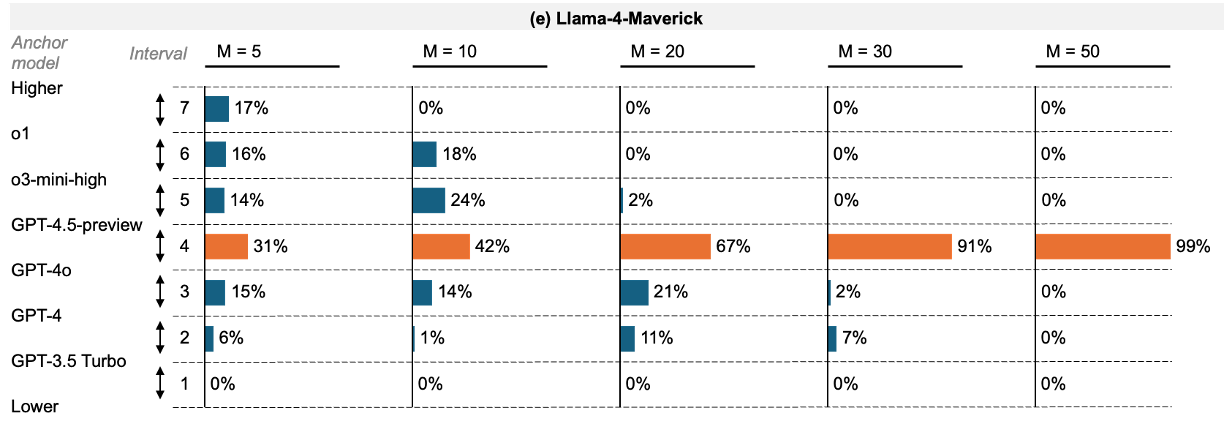}
  \vspace{2pt}
\end{subfigure}

\caption[Bayesian Model Ranking]{\textbf{Bayesian Probability Ranking Analysis}\\
\vspace{6pt}
\begin{minipage}{\textwidth}
Probability distributions of the test model's ranking relative to six anchor models across varying question counts ($M$).The anchor models partitioned the ranking space into seven mutually exclusive intervals, with probabilities quantifying the likelihood of the test model falling into each interval.
\end{minipage}
\vspace{-10pt}
}
\label{fig2}
\end{figure}

Bayesian inference positioned DeepSeek-R1 between o3-mini-high and o1 when evaluated with the full query set (Fig.~\ref{fig2}). Knowledge-intensive domains (sciences/humanities/arts) emerged as performance differentiators, where o1 demonstrated superior and more consistent performance relative to DeepSeek-R1. Robustness analysis confirmed stable interval identification down to M=20 questions, though probability sharpness progressively attenuated. When the number of questions is reduced to M=10 and less, the peak interval probability fell below 50\%, accompanied by increased ambiguity across adjacent categories.

\begin{figure}[htbp]
\centering
\includegraphics[width=1\textwidth]{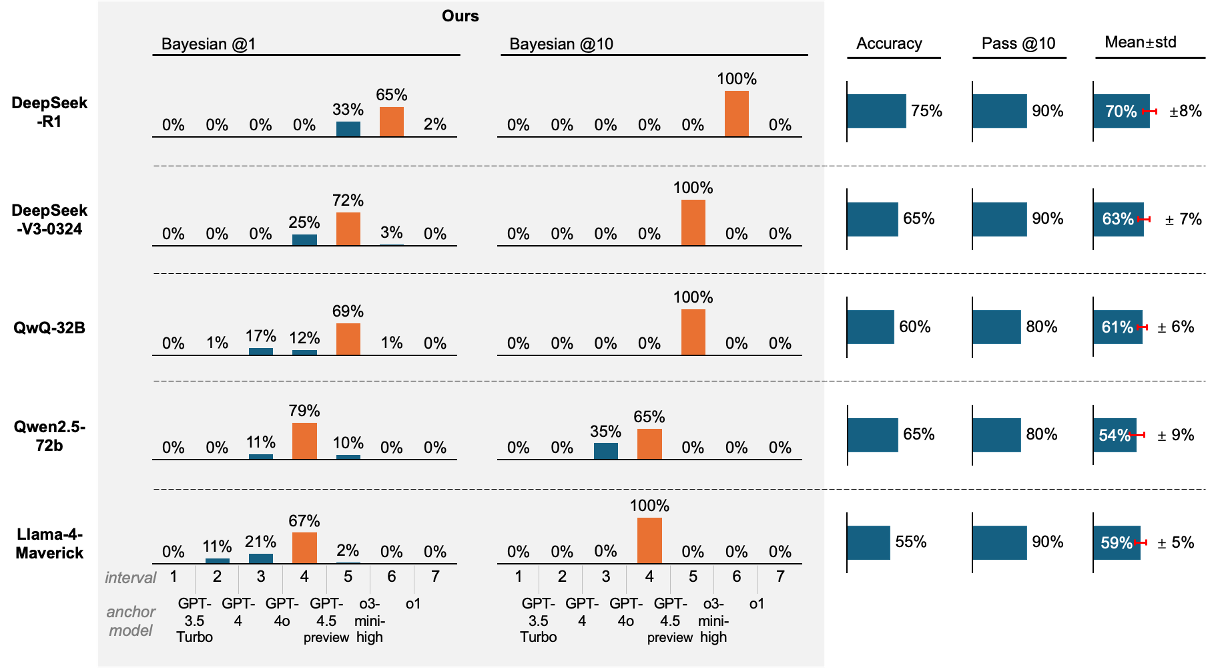}
\caption[Method comparison]{\textbf{Method comparison}\\
\vspace{4pt} 
\begin{minipage}{\textwidth}
Performance comparison between the proposed Bayesian approach and conventional evaluation metrics. Bayesian@1 and accuracy reporting are Single-trial evaluations, while Bayes@10, Pass@10 and Mean±std are aggregated from O=10 independent trials. All results are obtained from M=20 questions.
\end{minipage}
}
\label{fig3}
\end{figure}

Figure~\ref{fig2} also showed more validations with additional test models (DeepSeek-V3-0324, QwQ-32B, Qwen2.5-72B, Llama-4-Maverick) corroborating this pattern: full query sets enabled unambiguous capability localization, while $M=20$ retained at least $65\%$ confidence in most likely interval resolution. Sample size reduction to $M=10$ and below induced non-linear uncertainty propagation.

Figure~\ref{fig3} compared Bayesian evaluation metrics (single/multi-trial) against conventional reporting paradigms using $M=20$ questions, yielding four observations:

\begin{enumerate}
    \item Multi-trial Bayesian@10 demonstrated superior interval precision compared to Bayesian@1, while maintaining consistent category estimation boundaries across all models except Qwen2.5-72B. Notably, Qwen2.5-72B exhibited the largest standard error in mean+/std reporting, suggesting higher trial-to-trial variability.
    
    \item Accuracy reporting gives a single score for each model but does not provide any statistical interpretation on the reliability of the scores. Furthermore, it obscured performance hierarchies resolvable through Bayesian analysis, as evidenced by the 65\% accuracy achieved by both DeepSeek-V3-0324 and Qwen2.5-72B.
    
    \item Pass@10 aggregation inflated overall accuracy metrics while reducing discernibility between models. It does not provide any statistica interpretation either. In addition, Pass@10 suggested comparable performance for Llama-4-Maverick relative to DeepSeek-R1, which is the only evaluation metric suggesting such parity.  
    
    \item In traditional $\mathrm{mean}\pm\mathrm{std}$ reporting, DeepSeek-V3-0324, QwQ-32B, and Llama-4-Maverick all have means near 60\% with overlapping variances, suggesting comparable performance. However, Bayesian analysis indicated a categorical performance disparity, placing Llama-4-Maverick below the other two models despite their similar mean estimates.

\end{enumerate}

\section{Discussion}
Our systematic comparisons demonstrate Bayesian inference's superior discriminatory capacity over conventional evaluation methodologies through two  dimensions: statistical robustness under constrained query regimes, and resolution of performance ambiguities.

The Bayesian approach exhibits marked stability in discriminative performance when subjected to limited query budgets ($M=20$), outperforming conventional accuracy metrics that remain vulnerable to stochastic output variance while lacking rigorous statistical interpretation, particularly under low-sample conditions. While Pass@N frameworks establish meaningful performance ceilings, they exhibit critical limitations in resolving models with comparable best-case performances  - exemplified by Llama-4 showing a 10\% advantage in Pass@10 metrics yet underperforming QwQ-32B by one category with the Bayesian approach.

Mean±standard error reporting offered partial statistical transparency but suffered from resolution limitations in cross-model analyses. Notably, our Bayesian method resolved performance disparities between models with overlapping uncertainty intervals (QwQ-32B vs. Llama-4-Maverick), demonstrating superior discriminative power. The approach's incorporation of anchor model priors effectively addressed small-sample equivalence fallacies, where conventional metrics misattributed systematic capability differences to stochastic sampling noise. 

The proposed evaluation method enables progressive capability refinement through strategic anchor augmentation. For instance, incorporating QwQ-32B into the anchor ensemble could resolve its current equivalence with DeepSeek-V3-0324, demonstrating the method's adaptive capacity for step-by-step model discrimination through dynamic anchor integration.

While Item Response Theory (IRT) approaches like the Rasch model\citep{rasch1993probabilistic} have proven successful in educational assessment, their direct adaptation to LLM evaluation presents critical limitations. Both IRT and our approaches leverage anchor model priors when available, but diverge fundamentally in handling information uncertainty. Conventional IRT implementations assume Gaussian latent trait distributions across test populations - an assumption theoretically justified in human testing through central limit theorem applications but questionable with limited number of LLM models. In contrast, our Bayesian approach does not require such assumptions, accommodating non-Gaussian capability distributions while maintaining computational tractability. This relaxation of distributional constraints enhances flexibility compared to IRT-based methods, while maintaining granular, uncertainty-aware comparisons.

The evaluation method’s ability to deliver statistically calibrated rankings with about 20 queries aligns with real-world usage patterns. Experienced users naturally employ prior knowledge through iterative LLM interactions, instinctively comparing outputs against previously encountered models. Users also gain valuable insights from limited trials while quantifying uncertainty—e.g., "Model A is definitely better than GPT-3.5 but may be on par with GPT-4." Our methodology formalizes this cognitive process through explicit Bayesian inference. It systematically integrates priors derived from query set selection and anchor model selection, both of which reflect inherent human preference and experience. This methodology also recognizes expert judgment value - particularly critical when reliable prior information represents both scarce resource in the real-world and competitive advantage in cutting-edge LLM development.

\subsection{Limitations and Future Directions}
First, valid priors require high-quality query sets and representative anchors. While GPT-series models provided stable baselines, emerging architectures may disrupt assumed capability ranking orders. Automated anchor selection protocols and dynamic anchor updates warrant further investigation. 

Second, the factorization in Eqs. \ref{eq:3} assumes assumes conditional query independence -- a valid approximation in a well curated query set but may be invalid with semantically correlated queries. Future work should model question interdependencies or introduce non-correlated hierarchical priors (e.g., grouping questions by skill type) while examining independence assumptions across hierarchical levels.

Third, the binary scoring paradigm struggles with open-ended humanities questions. Extending the approach to multi-dimensional rubrics (e.g., correctness, creativity, coherence, empathy) would enhance versatility for complex tasks.

\section{Conclusion}
The study presents a method for evaluating LLMs by systematically incorporating measurable prior knowledge within Bayesian inference, thereby more effectively accommodating the inherent stochastic characteristics of these models. Notably, the proposed methodology demonstrates enhanced discriminative capability through prior knowledge integration even with limited sample sizes, while concurrently ensuring result stability in its analytical outcomes.

\setcitestyle{numbers}  
\bibliography{references}

\begin{appendix}

\clearpage 
\appendix
\renewcommand{\thesection}{Appendix \Alph{section}}
\section{Complete success rate distribution}
\label{Complete success rate distribution}

Complete success rate distribution of anchor models with 50 questions. Extreme probability values $\{0\%,100\%\}$ were modulated to $\{1\%,99\%\}$ to ensure numerical stability during subsequent computations.

\begin{figure}[htbp]
\centering
\begin{subfigure}{\textwidth}
  \centering
  \includegraphics[width=1\linewidth]{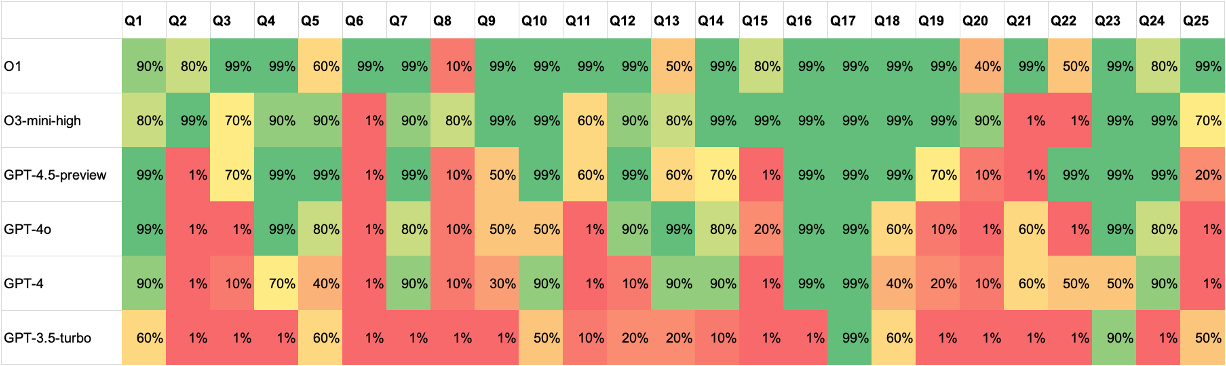}
  \vspace{0.2cm} 
\end{subfigure}
\vspace{0.1cm} 
\begin{subfigure}{\textwidth}
  \centering
  \includegraphics[width=1\linewidth]{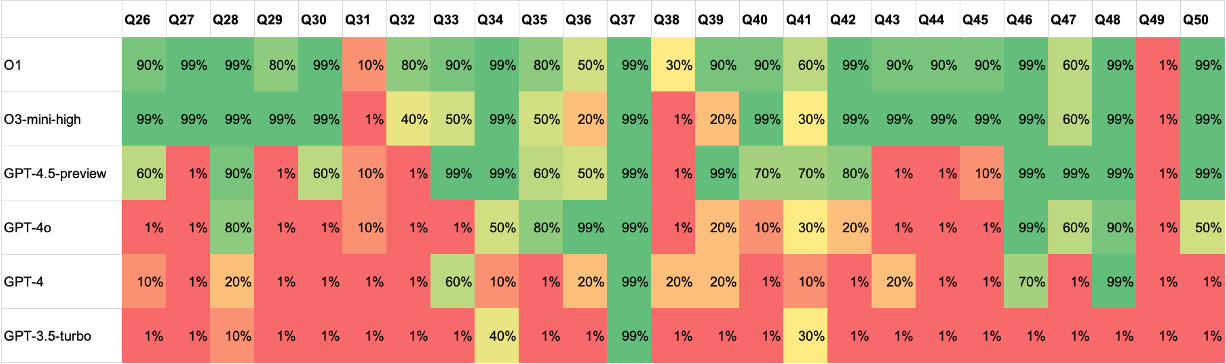}
  \vspace{0.2cm}
\end{subfigure}
\label{fig4}
\end{figure}

\clearpage 
\section{Complete Query Set}
\label{Complete Query Set}

When query set size was reduced to $M$, only the first $M$ questions were used from the complete 50-question-set.

{
\centering
\small
\setlength{\extrarowheight}{3pt} 
\begin{longtable}{|C{0.5cm}|L{9cm}|C{2cm}|C{2cm}|C{2cm}|}
\caption{\textbf{Complete query set with 50 questions}}\\
\hline
\textbf{NO} & \textbf{Question} & \textbf{Answer} & \textbf{Domains} & \textbf{Source} \\
\hline
\endfirsthead

\multicolumn{5}{c}%
{{\bfseries Table \thetable: Complete query set with 50 questions}} \\
\hline
\textbf{NO} & \textbf{Question} & \textbf{Answer} & \textbf{Domains} & \textbf{Source} \\
\hline
\endhead

\hline \multicolumn{5}{|r|}{{Continued on next page}} \\ \hline
\endfoot

\hline
\endlastfoot

1 & What is the large documentary called, that Sergey Apollinariyevich Grigoryev co-produced with Chinese film workers when he came to China in 1950?

\_\_\_A) Revolutionary Modern China.

\_\_\_B) Progressive China.

\_\_\_C) Liberated China.

\_\_\_D) Revolutionary New China.

\_\_\_E) Free China.

\_\_\_F) Modern China.

\_\_\_G) Revolutionary Free China.

\_\_\_H) New China.

\_\_\_I) Revolutionary Liberated China.

\_\_\_J) Revolutionary China & C & Arts & SuperGPQA \\
\hline

2 & The definition of an algorithm: The definition of a(n) is hypotenuse numbers (numbers that can be expressed as the square root of the sum of the squares of two non-zero integers). Given the input x\_list (a series of values): [55, 56, 57, 58, 59, 60, 61, 62, 63, 64], determine the corresponding output sequence y\_list. 

A) [118, 121, 124, 129, 131, 135, 136, 138, 143, 145]

B) [115, 118, 120, 127, 129, 134, 138, 139, 144, 146]

C) [114, 118, 121, 126, 129, 133, 137, 141, 142, 148]

D)[119, 120, 122, 123, 125, 130, 135, 136, 137, 140] 

E) [116, 119, 120, 121, 123, 126, 131, 133, 135, 138]

F) [122, 124, 125, 127, 131, 132, 138, 140, 141, 145]

G) [121, 122, 124, 126, 128, 132, 136, 139, 140, 142]

H) [124, 128, 130, 134, 137, 143, 147, 149, 150, 157]

I) [120, 125, 127, 131, 135, 136, 139, 141, 142, 149]

J) [117, 123, 126, 128, 133, 137, 139, 141, 143, 147] & D & Engineering & MMLU-Pro \\
\hline

3 & What is the major outcome of the reaction between 4,4-dimethylcyclopent-1-enol and bromine?

A. 2-bromo-4,4-dimethylcyclopentanone

B. (1R,2R)-1,2-dibromo-4,4-dimethylcyclopentanol

C. (1R,2S)-1,2-dibromo-4,4-dimethylcyclopentanol

D. 4-bromo-4,4-dimethylcyclopentanone & A & Chemistry & GPQA-Diamond \\
\hline

4 & What are the roots of $x^3-3x^2-10x+24$. List your answer as numbers separated by commas. & 2, -3, 4 & Mathematics & MATH \\
\hline

5 & Encryption Process Description

- Input Data:
- Plaintext: A string consisting solely of uppercase letters with no spaces or punctuation. 
- Output (Ciphertext): A string without punctuation.

- Setup: 
Use the following Multitap Code Table as a reference:

\begin{tabular}{|l|l|}
\hline
\textbf{Letter} & \textbf{Multitap Code} \\ \hline
A & 2\textsuperscript{1} \\ \hline
B & 2\textsuperscript{2} \\ \hline
C & 2\textsuperscript{3} \\ \hline
D & 3\textsuperscript{1} \\ \hline
E & 3\textsuperscript{2} \\ \hline
F & 3\textsuperscript{3} \\ \hline
G & 4\textsuperscript{1} \\ \hline
H & 4\textsuperscript{2} \\ \hline
I & 4\textsuperscript{3} \\ \hline
J & 5\textsuperscript{1} \\ \hline
K & 5\textsuperscript{2} \\ \hline
L & 5\textsuperscript{3} \\ \hline
M & 6\textsuperscript{1} \\ \hline
N & 6\textsuperscript{2} \\ \hline
O & 6\textsuperscript{3} \\ \hline
P & 7\textsuperscript{1} \\ \hline
Q & 7\textsuperscript{2} \\ \hline
R & 7\textsuperscript{3} \\ \hline
S & 7\textsuperscript{4} \\ \hline
T & 8\textsuperscript{1} \\ \hline
U & 8\textsuperscript{2} \\ \hline
V & 8\textsuperscript{3} \\ \hline
W & 9\textsuperscript{1} \\ \hline
X & 9\textsuperscript{2} \\ \hline
Y & 9\textsuperscript{3} \\ \hline
Z & 9\textsuperscript{4} \\ \hline
\end{tabular}

- Encryption Steps: 
For every character `p` in the plaintext:
1. Confirm that `p` is an uppercase letter and appears in the Multitap Code Table.
2. Substitute `p` with its corresponding Multitap Code from the table.

Decryption Process Description

- Input Data: 
- Ciphertext: A string devoid of punctuation.
- Output (Plaintext): A string of uppercase letters.

- Setup: 
The same Multitap Code Table is used as a reference.

- Decryption Steps: 
For every multitap code `c` in the ciphertext:
1. Confirm that `c` matches one of the codes in the table.
2. Replace `c` with the corresponding uppercase letter from the Multitap Code Table.

---

\#Question

Given the plaintext: `"UO"`, perform the encryption according to the above rules. 
Wrap your final encrypted output in double square brackets, e.g. `[[encrypted answer]]`. & [[8\textsuperscript{2}6\textsuperscript{3}]] & Reasoning & KOR-Bench \\

\hline

6 & When the furnace operates at normal temperature and the molten iron is flowing well, can the iron tap hole be ()? \newline 
A) Constricted\newline 
B) Smaller\newline 
C) No requirements\newline 
D) Elevated\newline 
E) Unchanged\newline 
F) Increased\newline 
G) Larger\newline 
H) Lowered\newline 
I) Expansive\newline 
J) Lessened & B & Materials Science & SuperGPQA \\
\hline

7 & When was the EP "Dirty Laundry" released by Adore Delano? \newline 
A) February 19, 2022\newline 
B) July 9, 2021\newline 
C) November 29, 2022\newline 
D) August 29, 2021\newline 
E) July 9, 2020\newline 
F) December 5, 2020\newline 
G) October 19, 2023\newline 
H) March 1, 2021\newline 
I) October 19, 2020\newline 
J) May 25, 2021 & B & Arts & MMLU-Pro \\
\hline

8 & Acetic acid is treated with bromine, pyridine, and acetic anhydride with heating, forming product 1.\newline 
1 is heated with ethanol and a small amount of sulfuric acid, forming product 2.\newline 
2 is treated with sodium cyanide, forming product 3.\newline 
3 is then treated with excess sodium hydride and 1,5-dibromopentane, forming final product 4.\newline 
how many distinct hydrogen signals will be observable in the 1H NMR spectrum of 4 (some of them maybe very close in chemical shift and thus not practically distinguishable, but the desired answer is the number of chemically distinct hydrogens)\newline 
A. 12\newline 
B. 8\newline 
C. 5\newline 
D. 10 & B & Chemistry & GPQA-Diamond \\
\hline

9 & What is the sum of all complex values of $a,$ such that the polynomial $x^4 + (a^2 - 1) x^2 + a^3$ has just two distinct complex roots. & 4 & Mathematics & MATH \\
\hline

10 & \textbf{Problem Statement}\newline 
You’re given a string composed of digits and lowercase letters, along with a starting character and a positive integer. Your task is to locate the character that lies a specified number of steps to the right of the starting character, moving one position per step and wrapping around to the beginning of the string when you reach the end.\newline 
At each step:\newline 
1. Move one position to the right in the string.\newline 
2. Decrement the step count by one.\newline 
3. Continue until the step count reaches zero.\newline 
You should report:\newline
- The **final** character found when the step count reaches zero.\newline
- The **initial** state as a pair [starting character, initial step count].\newline
- The **intermediate** states as a list of [character, remaining steps] pairs **after each move**, excluding the initial and final states.\newline 
\textbf{Input} \newline 
String: e6 \newline 
Command: ['6', '2'] & \{
"final": "6",
"init": [
"6",
"2"
],
"intermediate": [
[
"e",
"1"
]
]
\} & Computer Science & Procbench \\
\hline

11 & What is the university called where American music producer George Avakian began teaching in 1948? 

A) Columbia University.

B) University of Pennsylvania.

C) University of New England.

D) University of New York.

E) Harvard University.

F) University of New Hampshire.

G) Princeton University.

H) University of New Jersey.

I) New York University.

J) Yale University. & I & Arts & SuperGPQA \\
\hline

12 & What are the three theories about the origin of Proto-Malay language mentioned in the 'Encyclopedia of Malaysia: Early History'?

A) The Yunnan theory, the Seafarer theory, and the Borneo theory

B) The Yunnan theory, the Seafarer theory, and the Melanesian theory

C) The Yunnan theory, the Seafarer theory, and the Malay Archipelago theory

D) The Yunnan theory, the Seafarer theory, and the Philippine theory

E) The Yunnan theory, the Seafarer theory, and the Javanese theory

F) The Yunnan theory, the Seafarer theory, and the Micronesian theory

G) The Yunnan theory, the Seafarer theory, and the Taiwan theory

H) The Yunnan theory, the Seafarer theory, and the Sumatran theory

I) The Yunnan theory, the Seafarer theory, and the Polynesian theory

J) The Yunnan theory, the Seafarer theory, and the Indonesian theory & G & Humanities & MMLU-Pro \\
\hline

13 & A textile dye containing an extensively conjugated pi-electrons emits light with energy of 2.3393 eV. What color of light is absorbed by the organic compound?

A. Violet

B. Yellow

C. Red

D. Blue & C & Chemistry & GPQA-Diamond \\
\hline

14 & Consider four integers where each leaves a different remainder when divided by 6. If multiplying these integers together results in a value $ Y $ that is not divisible by 6, determine the remainder obtained when $ Y $ is divided by 6. & 4 & Mathematics & MATH \\
\hline

15 & \#Rule: \newline
- Literal: A propositional variable and its negation together are called literals. \newline
- Complement: For any literal L, its complement is written as L'. That is, if L is p then L' is $\lnot$p; if L is $\lnot$p then L' is p. \newline
- Resolution: \newline
Suppose you have two simple disjunctive clauses: \newline
- C1 = C3 $\lor$ L \newline
- C2 = C4 $\lor$ L' \newline
Then, C1 and C2 can be resolved, and the operation is defined as dispel(C1, C2) = C3 $\lor$ C4. If the result is empty, then dispel(C1, C2) = 0. \newline
- Resolution Algorithm: \newline
1. Input: A conjunctive normal form S. \newline
2. Output: Output "Plausible" if S has a satisfying assignment; otherwise, output "Implausible." \newline
3. Procedure: \newline
- Initialization: \newline
- Let S0 and S2 be empty sets. \newline
- Let S1 be the set of all simple disjunctive clauses in S. \newline
- Resolution with S0 and S1: \newline
- For each clause C1 in S0 and each clause C2 in S1, if they can be resolved, compute C = dispel(C1, C2). \newline
- If C = 0, output "Implausible" and stop. \newline
- If C is new (i.e., not present in S0 or S1), add C to S2. \newline
- Resolution within S1: \newline
- For every pair of clauses C1 and C2 in S1, if they can be resolved, compute C = dispel(C1, C2). \newline
- If C = 0, output "Implausible" and stop. \newline
- If C is new, add C to S2. \newline
- Check S2: \newline
- If S2 is empty, output "Plausible" and end. \newline
- Otherwise, move S1 into S0, set S1 equal to S2, clear S2, and repeat from the earlier resolution steps. \newline
\#Question: \newline
Given the clauses \newline
- C1 = $\lnot$p $\lor$ q $\lor$ r \newline
- C2 = p $\lor$ $\lnot$r $\lor$ $\lnot$s \newline
calculate dispel(C1, C2). \newline
Provide the answer in the format `[[]]`, or `[[];[];\ldots]` if multiple answers exist.& 
\makecell[l]{
{[}{[}q$\lor$r$\lor$$\neg$r$\lor$$\neg$s{]}; \\
{[}$\neg$p$\lor$q$\lor$p$\lor$$\neg$s{]}
} & Reasoning & KOR-Bench \\
\hline

16 & The sequence of shell plating starts from the flat keel's K strake and extends toward the \_\_\_\_\_, with each strake labeled sequentially as A, B, C, D, etc., until reaching the \_\_\_\_\_. \newline Options: \newline A) Upwards / adjacent strakes \newline B) Port and starboard directions / deck strakes \newline C) Forward direction / subsequent strakes \newline D) The port and starboard sides / side plates & B & Engineering & SuperGPQA \\
\hline

17 & Very Low Birth Weight (VLBW) has the meaning of \_\_\_. \newline \_\_\_\_A) Less than 1500g within 1 hour of birth \newline \_\_\_\_B) Less than 1000g within 24 hours of birth \newline \_\_\_\_C) Less than 1500g after 24 hours of birth \newline \_\_\_\_D) More than 1500g after 24 hours of birth \newline \_\_\_\_E) More than 1700g within 24 hours of birth \newline \_\_\_\_F) More than 2000g within 1 hour of birth \newline \_\_\_\_G) Less than 1300g after 30 minutes of birth \newline \_\_\_\_H) Less than 1200g after 1 hour of birth \newline \_\_\_\_I) More than 1000g within 1 hour of birth \newline \_\_\_\_J) More than 1500g within 1 hour of birth & A & Medical Science & MMLU-Pro \\
\hline

18 & Which particle is not associated with a spontaneously-broken symmetry? \newline A. Pion \newline B. Magnon \newline C. Phonon \newline D. Skyrmion & D & Physics & GPQA-Diamond \\
\hline

19 & $G$ and $H$ are the centroid and orthocenter of triangle $ABC,$ respectively. Let $F$ be the midpoint of $\overline{GH}.$ Show $AF^2 + BF^2 + CF^2$ in terms of the side lengths $a,$ $b,$ $c$ and circumradius $R$ of triangle $ABC.$ & $3R^2$ & Mathematics & MATH \\
\hline

20 & \textbf{Rule:} \newline In a simple conjunctive form (or a simple disjunctive form) that contains n propositional variables, if each propositional variable and its negation appear exactly once, and the propositional variables or their negations are arranged in ascending order of subscripts or in lexicographical order, such a form is called a paired conjunctive term (or paired disjunctive term). \newline If the true assignment of a paired conjunctive term corresponds to a binary number equal to hexadecimal number i, this paired conjunctive term is denoted as mi (with lowercase m). For example, the true assignment of p$\land$q is 11, and the binary number 11 corresponds to hexadecimal 3, which is denoted as m3. \newline Similarly, if the false assignment of a paired disjunctive term corresponds to a binary number equal to hexadecimal number i, this paired disjunctive term is denoted as Mi (with uppercase M). For instance, the false assignment of $\neg$p$\lor$$\neg$q$\lor$$\neg$r is 111, and the binary number 111 (hexadecimal 7) is denoted as M7. \newline The disjunctive normal form (or conjunctive normal form) composed of all paired conjunctive terms (or paired disjunctive terms) is called the principal disjunctive normal form (or principal conjunctive normal form). \newline Given a formula A containing n propositional variables: \newline - If the principal disjunctive normal form of A includes all $2^n$ paired conjunctive terms, then A is a tautology. \newline - If the principal disjunctive normal form of A includes no paired conjunctive terms, then A is a contradiction. \newline - If the principal disjunctive normal form of A includes m0, then A is a basic formula. \newline - If the indices i of the paired conjunctive terms in the principal disjunctive normal form of A are all even, then A is an all-even formula. \newline - If the indices i of the paired conjunctive terms in the principal disjunctive normal form of A are all odd, then A is an all-odd formula. \newline \textbf{Question:} \newline Given that formula A contains 4 propositional variables, what should it be denoted as if it is both a tautology and a basic form? \newline Provide your answer wrapped in double square brackets. & 
\begin{minipage}[t]{3.5cm}
\vspace{-0.5em}
{[[}m0$\lor$m1$\lor$m2$\lor$\newline
m3$\lor$m4$\lor$m5$\lor$\newline
m6$\lor$m7$\lor$m8$\lor$\newline
m9$\lor$mA$\lor$mB$\lor$\newline
mC$\lor$mD$\lor$mE$\lor$\newline
mF{]]}
\end{minipage} & Reasoning & KOR-Bench \\
\hline

21 & When was "Discussion sur l'évolution de l'univers" by Georges Lemaitre published?\newline 
A) 1931\newline 
B) 1927\newline 
C) 1933\newline 
D) 1935\newline 
E) 1937\newline 
F) 1929\newline 
G) 1932\newline 
H) 1930\newline 
I) 1936\newline 
J) 1934 & C & Physics & SuperGPQA \\
\hline

22 & Which weather condition is characterized by the presence of a layer of high, thin clouds often preceding a warm front?\newline 
A) Cumulus clouds\newline 
B) Cumulonimbus clouds\newline 
C) Altocumulus clouds\newline 
D) Cirrus clouds\newline 
E) Stratocumulus clouds\newline 
F) Stratus clouds\newline 
G) Altostratus clouds\newline 
H) Cirrocumulus clouds\newline 
I) Cirrostratus clouds\newline 
J) Nimbostratus clouds & D & Earth Science & SuperGPQA \\
\hline

23 & The 'Encyclopedia of Malaysia: Early History' mentioned which three theories about the origin of Proto-Malay language?\newline 
A) The Yunnan theory, the Seafarer theory, and the Taiwan theory\newline 
B) The Yunnan theory, the Seafarer theory, and the Melanesian theory\newline 
C) The Yunnan theory, the Seafarer theory, and the Malay Archipelago theory\newline 
D) The Yunnan theory, the Seafarer theory, and the Philippine theory\newline 
E) The Yunnan theory, the Seafarer theory, and the Borneo theory\newline 
F) The Yunnan theory, the Seafarer theory, and the Javanese theory\newline 
G) The Yunnan theory, the Seafarer theory, and the Micronesian theory\newline 
H) The Yunnan theory, the Seafarer theory, and the Sumatran theory\newline 
I) The Yunnan theory, the Seafarer theory, and the Polynesian theory\newline 
J) The Yunnan theory, the Seafarer theory, and the Indonesian theory & A & Humanities & MMLU-Pro \\
\hline

24 & What are the building blocks of a pattern training method?\newline 
A) Core elements, inspection standards, assessment criteria, training techniques.\newline 
B) Key pattern, setup procedures, testing benchmarks, educational systems.\newline 
C) Target model, inspection means, evaluation criteria, training methods.\newline 
D) Pattern design, implementation tools, review standards, training models.\newline 
E) Principal components, examination vehicles, judging norms, learning techniques.\newline 
F) Primary algorithm, review guidelines, execution tactics, learning processes.\newline 
G) Basic units, examination means, evaluation frameworks, instructional methods.\newline 
H) Essential components, inspection measures, validation guidelines, learning models.\newline 
I) Basic algorithms, inspection tools, evaluation metrics, pattern methods.\newline 
J) Foundational structure, evaluation tools, training regimens, feedback mechanisms. & C & Humanities & MMLU-Pro \\
\hline

25 & 1-bromobenzene-2-d is treated with NaNH2 in condensed ammonia solvent. This reaction contains how many possible organic products?\newline 
A. 2\newline 
B. 1\newline 
C. 3\newline 
D. 4 & C & Chemistry & GPQA-Diamond \\
\hline

26 & In a specific region of the sky, astronomers have observed that the number of stars varies with parallax as $1/\text{plx}^5$. How does the number of stars in that region of the sky change with distance (per unit range of distance, $r$)?\newline 
A. $\sim r^2$\newline 
B. $\sim r^3$\newline 
C. $\sim r^4$\newline 
D. $\sim r^5$ & B & Physics & GPQA-Diamond \\
\hline

27 & List all possible values of the determinant of\newline 
$\begin{pmatrix} \sec^2 x & 1 & 1 \\ \cos^2 x & \cos^2 x & \csc^2 x \\ 1 & \cos^2 x & \cot^2 x \end{pmatrix}$,\newline 
as $x$ ranges over all real numbers (where the determinant is defined). & (0,1) & Mathematics & MATH \\
\hline

28 & Felix generates a two-digit integer by rolling a six-sided dice twice. The result of his first roll is the tens digit, and the result of his second roll is the ones digit. Find the probability that the resulting integer is divisible by 8? Express your answer as a common fraction. & $\dfrac{5}{36}$ & Mathematics & MATH \\
\hline

29 & \textbf{Rules}:\newline 
1. The game begins with an initial word and identifies a target word at the end.\newline 
2. You may change only one letter at each step, with every resulting intermediate word being valid.\newline 
3. Transform the starting word into the target word using the fewest number of steps possible.\newline 
4. The puzzle provides the starting and target words. Your task is to determine the minimum number of transformations required.\newline 
\textbf{Task}:\newline 
Convert the word "HEAD" into "TALE". Express the minimum number of steps needed in double square brackets. For instance, if it requires 3 steps, write your answer as [[3]]. & [[5]] & Reasoning & KOR-Bench \\
\hline

30 & \textbf{Example Puzzle}:\newline 
"example\_puzzle": "There are three houses arranged in a row, numbered 1 to 3 from left to right as seen from the opposite side of the street. Each house is inhabited by a different person, and every house is linked with a unique attribute in two categories: the resident's name and their preferred drink. The available names are Peter, Eric, and Arnold, while the drinks are tea, water, and milk. The clues provided are as follows:\newline 
1. Peter occupies the second house.\newline 
2. Arnold lives immediately to the left of the person who drinks only water.\newline 
3. The person who drinks only water is immediately to the left of the person whose favorite drink is milk.\newline 
Please explain your reasoning and provide your final answer using the JSON format below:\newline 
\{ "reasoning": "",
 "solution": 
  \{ 
 "House 1": 
\{ "Name": "",
 "Drink": ""\}, 
 "House 2": 
\{  "Name": "",
 "Drink": ""\}, 
 "House 3":  
\{ "Name": "",
 "Drink": ""\}
\}\}", 
\newline "example\_puzzle\_answer": "\{"reasoning": "Given Clue 1, Peter must be in House 2. Clue 2 indicates that Arnold is immediately to the left of the person who drinks only water; since the third house cannot have anyone to its left, Arnold must reside in House 1. Therefore, the resident in House 2 (Peter) is the one who drinks water, leaving House 3 for Eric. Clue 3 then implies that Eric's favorite drink is milk, so by elimination Arnold must prefer tea.", "solution": \{"House 1": \{"Name": "Arnold", "Drink": "tea"\}, "House 2": \{"Name": "Peter", "Drink": "water"\}, "House 3": \{"Name": "Eric", "Drink": "milk"\}\}\}", 
\newline \textbf{Puzzle\_to\_solve}: 
\newline"Imagine five houses lined up from left to right, numbered 1 to 5 as seen from the other side of the street. Each house is occupied by a different individual, and every house features a distinct characteristic in the following four categories:\newline 
- Resident's name: Arnold, Peter, Bob, Alice, Eric\newline 
- Car model: toyota camry, tesla model 3, bmw 3 series, honda civic, ford f150\newline 
- Favorite color: blue, red, green, white, yellow\newline 
- Mother's name: Janelle, Penny, Holly, Kailyn, Aniya\newline 
The clues are provided below:\newline 
1. The person who drives a Honda Civic is the one whose favorite color is yellow.\newline 
2. The individual whose favorite color is red drives a Tesla Model 3.\newline 
3. The owner of the BMW 3 Series does not reside in the fourth house.\newline 
4. The person whose mother is named Aniya is the one who prefers blue.\newline 
5. Eric's favorite color is green.\newline 
6. The resident with red as their favorite color lives somewhere to the left of the owner of a Ford F-150.\newline 
7. Alice is immediately to the left of Eric.\newline 
8. The resident whose mother is Holly lives in the first house.\newline 
9. Arnold has white as his favorite color.\newline 
10. The person whose mother is Janelle loves white.\newline 
11. The resident whose mother is Kailyn is Alice.\newline 
12. Arnold resides somewhere to the left of Peter.\newline 
13. Eric is the owner of the BMW 3 Series.\newline 
14. Bob lives adjacent to the house where the owner of the Ford F-150 resides.\newline 
Please detail your reasoning process and provide your final solution using the JSON structure below: \{ "reasoning": "", "solution": \{ "House 1": \{ "Name": "", "CarModel": "", "Color": "", "Mother": "" \}, "House 2": \{ "Name": "", "CarModel": "", "Color": "", "Mother": "" \}, "House 3": \{ "Name": "", "CarModel": "", "Color": "", "Mother": "" \}, "House 4": \{ "Name": "", "CarModel": "", "Color": "", "Mother": "" \}, "House 5": \{ "Name": "", "CarModel": "", "Color": "", "Mother": "" \} \}\}" & \{"header": ["House", "Name", "CarModel", "Color", "Mother"], "rows": [["1", "Bob", "tesla model 3", "red", "Holly"], ["2", "Arnold", "ford f150", "white", "Janelle"], ["3", "Peter", "toyota camry", "blue", "Aniya"], ["4", "Alice", "honda civic", "yellow", "Kailyn"], ["5", "Eric", "bmw 3 series", "green", "Penny"]]\} & Reasoning & ZebraLogic \\
\hline

31 & Among the listed grades, which gray cast iron has a matrix structure combining ferrite and pearlite?\newline A) HT450\newline B) HT350\newline C) HT250\newline D) HT550\newline E) HT300\newline F) HT150\newline G) HT100\newline H) HT200\newline I) HT500\newline J) HT400 & F & Materials Science & SuperGPQA \\
\hline

32 & The structural basis for the 'after-discharge' effect in reflex activities is what type of connection between central neurons?\newline A) Divergent\newline B) Parallel\newline C) Ring\newline D) Spiral\newline E) Loop\newline F) Chain\newline G) Circular\newline H) Convergent & C & Biology & MMLU-Pro \\
\hline

33 & Which of the following is correct when things are in quantitative change?\newline A) Modifications in the volume of materials\newline B) Changes in the quantity of things\newline C) Adjustments in the count of objects\newline D) Fluctuations in the level of elements\newline E) Modifications in the count of things\newline F) Alterations in the tally of materials\newline G) Transformations in the tally of entities\newline H) Shifts in the amount of items\newline I) Alterations in the number of objects\newline J) Shifts in the number of items & B & Humanities & SuperGPQA \\
\hline

34 & What are the delta deposits actually?\newline A) Tropical zones\newline B) Maritime and terrestrial mixed\newline C) Volcanic material\newline D) Marine facies\newline E) Fluvial sediment\newline F) Coastal deposition\newline G) Glacial sediment\newline H) Continental\newline I) Desert areas\newline J) Oceanic basins & B & Earth Science & SuperGPQA \\
\hline

35 & Just before Augustine George Masih was appointed as a judge of the Supreme Court of India,what was his position?\newline A) Former chief justice of the Rajasthan High Court\newline B) Former chief justice of the Himachal Pradesh High Court\newline C) Former chief justice of the Jharkhand High Court\newline D) Former chief justice of the Chhattisgarh High Court\newline E) Former chief justice of the Delhi High Court\newline F) Former chief justice of the Calcutta High Court\newline G) Former chief justice of the Allahabad High Court\newline H) Former chief justice of the Bombay High Court\newline I) Former chief justice of the Punjab and Haryana High Court\newline J) Former chief justice of the Madras High Court & A & Humanities & MMLU-Pro \\
\hline

36 & Which of the following are the types of metamorphic reactions?\newline A) Solids with moderate temperature\newline B) Solids at low temperature\newline C) Fluids with high temperature\newline D) Gases under moderate temperature\newline E) Gases with moderate temperature\newline F) Gases at high temperature\newline G) Fluids with low temperature\newline H) Solids with high temperature\newline I) Fluids at low temperature\newline J) Gases at low temperature & C & Earth Science & MMLU-Pro \\
\hline

37 & What type of natural disaster data is primarily collected by dropsondes?\newline A) Tropical waves\newline B) Tropical cyclones\newline C) Hurricanes\newline D) Severe thunderstorms\newline E) Monsoons\newline F) Tropical disturbances\newline G) Typhoons\newline H) Tropical depressions\newline I) Cyclonic storms\newline J) Tropical storms & B & Earth Science & MMLU-Pro \\
\hline

38 & Which is the earliest modern Chinese play?\newline A) The Big Event\newline B) Shrew\newline C) A Night in the Coffee Shop\newline D) Zhao Yanwang & D & Arts & MMLU-Pro \\
\hline

39 & Very large number of neutrinos produced by the Sun reach the Earth. Let us assume that, hypothetically, the pp-III branch suddenly stopped in the core of the Sun about 8 and a half minutes ago, while all other reactions remained as they were. What would be the approximate ratio of the flux between two bands of neutrino energies of 700-800 KeV (band 1) and 800-900 keV (band 2). Flux (band 1) / flux (band 2) is:\newline A) 10\newline B) 1\newline C) 0.01 (10\textsuperscript{-2})\newline D) 0.1 (10\textsuperscript{-1}) & C & Physics & GPQA-Diamond \\
\hline

40 & Substance X, known for incorporating a heavier isotope of one of its constituent elements, reacts violently with liquid Y with the release of a gas W whose molecule contains the same number of neutrons and protons, and a precipitate G forms, which, when heated, releases B. The melting point of B (under normal conditions) is very close to 277 K. The product of the reaction of a certain keto acid with the substance X contains 2 atoms of oxygen. The substance X and especially its very close analog is used as a reagent in organic chemistry. Calculate the cumulative atomic masses of the lightest and heaviest elements present within Substance X, considering that if multiple instances of an element exist, the masses of all the heavier and lighter isotopes must be summed.\newline A) 29\newline B) 32\newline C) 25\newline D) 35 & D & Chemistry & GPQA-Diamond \\
\hline

41 & An atomic nucleus of mass M is at rest with rest-mass energy of 300 GeV. A spontaneous fission occurs in which it splits into two fragments (and nothing else), such that one fragment is 2 times more massive than the other (in terms of their rest-masses). The sum of rest-masses of the two fragments is 99\% of the initial mass M.\newline Kinetic energy of the more massive fragment is T1. What is the difference between the (correct) T1 value, and T1 value calculated using classical (non-relativistic) approximation?\newline (Ignore the electrons.)\newline A. 5 MeV.\newline B. 10 MeV.\newline C. 2 MeV.\newline D. 20 MeV. & A & Physics & GPQA-Diamond \\
\hline

42 & An equimolar mixture of salts A and B weighing 7.20 g was heated to 200°C without air. In this case, only a mixture of gases was formed, which, without cooling, was passed successively through tubes filled with anhydrous Mg(ClO4)2 (№1), Ca(OH)2 (№2) solution, and red-hot copper (№3). Results: the weights of tubes №1 and №3 increased by 3.60 g and 0.80 g respectively (CuO formed in tube №3). The weight of the second tube has not changed. As a result, only 2.24 liters of gas C (standard temperature and pressure) remained.\newline Find the total number of all atoms in salts A and B.\newline A. 15\newline B. 17\newline C. 19\newline D. 13 & B & Chemistry & GPQA-Diamond \\
\hline

43 & Two circles of diameter 2 are centered at (4,0) and (-4,0). How many circles are tangent to both of the given circles and also pass through the point (0,5)? & 4 & Mathematics & MATH \\
\hline

44 & Let $z_1$, $z_2$, $z_3$, and $z_4$ be the four distinct complex solutions of equation\newline 
$z^4 - 6z^2 + 8z + 1 = -4(z^3 - z + 2)i$.\newline 
What is the sum of the six pairwise distances between $z_1$, $z_2$, $z_3$, and $z_4$ in the complex plane. & $6\sqrt{3}+6$ & Mathematics & MATH \\
\hline

45 & Real number $x$ satisfies\newline 
$3x + \frac{1}{2x} = 3$.\newline 
What is\newline 
$64x^6 + \frac{1}{729x^6}$. & $\frac{416}{27}$ & Mathematics & MATH \\
\hline

46 & The school has 360 students. 15 take calculus, physics, and chemistry, and 15 don't take any of them. 180 take calculus. Twice as many students take chemistry as take physics. 75 take both calculus and chemistry, and 75 take both physics and chemistry. Only 30 students take both physics and calculus.\newline Find the number of students that take physics? & 110 & Mathematics & MATH \\
\hline

47 & \textbf{Task}\newline Given a string, split it at specified positions provided as a list of index pairs (i, j) using 0-based indexing. Only count the characters originally present in the text; do not include the delimiters introduced during the splitting process.\newline Starting with the first index pair, select the i-th substring from the current list and break it at the j-th character. This produces two new substrings: one containing the characters from index 0 to j-1, and the other containing the rest. Continue to process each index pair in order.\newline Return a 2D array that captures the list of substrings immediately after each individual split operation. The output should not include the initial unsplit list or the final result.\newline \textbf{Question}\newline String:\newline zrzbxzupqmwpokqpyuisg\newline Index Pairs:\newline (0, 9), (0, 4), (0, 1), (3, 1), (1, 2)  & 
"final": ["z","rz","b",\newline 
"xzupq","m",\newline 
"wpokqpyuisg"]
 & Reasoning & Procbench \\
\hline

48 & Intensional definitions classify how terms are defined by their essential attributes. Below are categories of intensional definitions:\newline 
1. $\diamond$ Definition: Defines a concept by identifying its broader category (genus) and distinguishing traits (differentia), structured as "term = differentia + genus."\newline 
2. $\star$ Definition: Specifies the origin or source of the entity described by the concept as the differentia.\newline 
3. $\dagger$ Definition: Uses the entity's functional role or purpose as the differentia.\newline 
4. $\circ$ Definition: Defines a concept through its relational attributes to other entities.\newline 
5. $\bullet$ Definition: Describes operational procedures or processes to delineate the term.\newline 
6. $\ast$ Definition: Employs logical expressions, often for relational concepts outside genus-differentia frameworks.\newline 
\textbf{Question}:\newline 
"Nuclear power, also termed atomic power, is the energy generated from processes that alter the composition of an atom's nucleus."\newline 
Which intensional definition category does this statement exemplify?\newline 
A. $\diamond$ Definition\newline 
B. $\dagger$ Definition\newline 
C. $\circ$ Definition\newline 
D. $\star$ Definition\newline 
E. $\bullet$ Definition\newline 
F. $\ast$ Definition & [D] & Reasoning & KOR-Bench \\
\hline

49 & \textbf{Rule}:\newline The Eagles of Manwë are extraordinary beings capable of traveling faster than any other method of transportation across Middle-earth.\newline The Red Book of Westmarch preserves the history and lore of Middle-earth, containing stories of hobbits, powerful rings, and the dark lord.\newline In Middle-earth, the Ring-bearer held the responsibility of guarding the One Ring, the most dangerous and precious artifact.\newline The One Ring's power was so immense that it could shape the fate of entire civilizations and realms in Middle-earth.\newline Knowing the Elven language, one of the most ancient and wise tongues, would be crucial for resolving conflicts between the diverse peoples of Middle-earth.\newline The Palantíri were mystical seeing-stones enabling long-distance communication, though with certain dangers and limitations.\newline The Arkenstone, revered as the "Heart of the Mountain," symbolized unity and leadership among the Dwarves of Erebor.\newline The archives of Minas Tirith contain an extensive collection of records on Gondor's history and alliances.\newline Helm's Deep is famous for its strong defenses, making it a nearly impenetrable fortress in Middle-earth.\newline Gandalf's wisdom was often sought after for guidance in crucial decisions that shaped Middle-earth's fate.\newline Dwarven craftsmanship is unmatched, particularly in rebuilding and fortifying cities.\newline The Mirror of Galadriel allows visions of distant events, including potential future occurrences, such as weather patterns over Middle-earth.\newline The Council of Elrond facilitated the alliance of Elves, Men, Dwarves, and Hobbits to combat Sauron's growing threat.\newline The Ents of Fangorn Forest protect the ancient trees and their inhabitants from harm.\newline Minas Tirith's archives are a rich repository of historical records, chronicling the deeds and legacies of Men's kingdoms.\newline Galadriel's wisdom, with her profound insight and discernment, is a powerful tool for distinguishing truth from deception in Middle-earth.\newline The Shire's fertile soil ensures bountiful harvests and sustains the hobbits' food supplies.\newline The Rangers, masters of survival and combat, ensure safe passage for travelers in the perilous wilderness of Middle-earth.\newline Ent-draughts, with their healing properties, are a unique solution to water scarcity in Middle-earth's forests.\newline \textbf{Question}:\newline Which skill is crucial for a mediator when resolving conflicts?\newline Options:\newline A. Legal knowledge.\newline B. Mastery of the Elven tongue.\newline C. Communication skills.\newline D. Knowledge of the Palantíri.\newline Provide your final answer as a single uppercase letter representing the option (A, B, C, or D) and wrap it in double square brackets, like this: [[A]]. & [[B]] & Reasoning & KOR-Bench \\
\hline

50 & \textbf{Rule}:\newline Define a binary operator "※" for integers with the following properties:\newline 1. If a is a multiple of b, then a※b = a/b + 2.\newline 2. If b is a multiple of a, then a※b = b/a + 2.\newline 3. If neither number is a multiple of the other, then a※b = 24.\newline \textbf{Problem}:\newline Using the rules above, evaluate the expression 25※5※14 sequentially.\newline Your answer should be a single number, formatted within double square brackets (for example: [[your answer]]). & [[4]] & Reasoning & KOR-Bench \\
\hline

\end{longtable}
}

\end{appendix}

\end{document}